\documentclass[twoside,letterpaper,journal]{IEEEtran}
\usepackage[utf8]{inputenc} 
\usepackage[T1]{fontenc}   
\usepackage{subcaption}
\usepackage{booktabs} 
\usepackage{amsmath,amsfonts} 
\usepackage{array}
\usepackage{textcomp}
\usepackage{stfloats}
\usepackage{url}
\usepackage{verbatim}
\usepackage{graphicx}
\usepackage{multicol}
\usepackage{siunitx}
\usepackage{tikz}
\usepackage{color}   
\usepackage{xcolor}   
\usepackage[linesnumbered,ruled,vlined]{algorithm2e}
\usepackage{multirow} 
\usepackage{cite}  
\usepackage{hyperref} 
\hypersetup{
    colorlinks=true,    
    linkcolor=blue,     
    citecolor=green,    
    filecolor=magenta,  
    urlcolor=cyan       
}

\usepackage{cleveref} 
\crefname{table}{Table}{Tables}
\crefname{figure}{Fig.}{Fig.}
\crefname{equation}{Eq.}{Eq.}
\crefname{section}{Section}{Sections}

\newcolumntype{P}[1]{>{\centering\arraybackslash}p{#1}}
\newcolumntype{C}[1]{>{\centering\arraybackslash}m{#1}}

\newcommand*{\rom}[1]{\expandafter\@slowromancap\romannumeral #1@}

\SetCommentSty{mycommfont}
\SetKwInput{KwInput}{Input}            
\SetKwInput{KwOutput}{Output}

\let\oldnl\nl
\newcommand{\nonl}{\renewcommand{\nl}{\let\nl\oldnl}}
 
\begin{document}

\title{EgoNav: Bridging Learned Waypoints and Geometry-Aware Local Control for Robust Indoor Navigation}

\author{
Jing Wang,
Shiqi~Zhao,
Hairong~Qu,
Peng~Yin\textsuperscript{*} 
 \thanks{Manuscript received: March 28, 2026; Revised: June 21, 2026; Accepted: July 24, 2026.}%
 \thanks{This paper was recommended for publication by Editor Aniket Bera upon evaluation of the Associate Editor and Reviewers comments.}%
\thanks{Peng Yin, Jing Wang, Shiqi Zhao and Hairong Qu are with the City University of Hong Kong, Hong Kong, China (pengyin@cityu.edu.hk; jingwan275-c@my.cityu.edu.hk; shiqizhao6-c@my.cityu.edu.hk; hairongqu2-c@my.cityu.edu.hk);}
\thanks{\textsuperscript{*}Corresponding author: Peng Yin.} 
\thanks{Digital Object Identifier (DOI): see top of this page.}%
}


\markboth{IEEE ROBOTICS AND AUTOMATION LETTERS. PREPRINT VERSION. ACCEPTED AUGUST, 2026}%
{Wang \MakeLowercase{\textit{et al.}}: EgoNav: Bridging Learned Waypoints and Geometry-Aware Local Control for Robust Indoor Navigation}

\maketitle

\begin{abstract}

Image-goal navigation using lightweight topological maps is a
practical paradigm for indoor robot deployment: the map
requires only geotagged images, and localization relies on
visual matching rather than precise pose estimation. However,
learned waypoint predictors can produce targets that violate
geometric constraints or deviate from the global path.
Executing these waypoints safely further requires a local
planner capable of collision avoidance, yet existing systems
either lack one or rely on fixed parameters that cannot adapt
to confined spaces. 
To address these limitations while retaining the navigational
intuition of the learned predictor, we present EgoNav, a
hierarchical system that implements this idea by generating
candidates from semantically segmented traversable regions
and scoring them alongside the learned waypoint for geometric
safety, directional coherence, and fidelity to the learned
prior. An adaptive local path planner then executes the
refined waypoint with parameters modulated based on the
refinement outcome. 
Experiments in Habitat-sim and on
a physical humanoid robot show that EgoNav consistently
outperforms contemporary baselines in both success rate and
path efficiency.
\end{abstract}


\section{Introduction}

Robust indoor navigation requires both global path awareness
and reliable local control. 
Existing systems that rely on precise metric maps
and accurate global localization remain fragile under
localization drift and environmental
change~\cite{planner:fast-planner, planner:ego-planner}.
Mapless alternatives avoid this dependency but introduce
other limitations. Object-Goal
Navigation~\cite{poni, aligning} resorts to inefficient
exploration over longer distances, while Vision-Language
Navigation~\cite{vln:navgpt, vln:towards} further requires
detailed step-by-step instructions that are impractical to
obtain at scale.
A practical navigation system should provide global path
awareness without requiring precise metric infrastructure.

\begin{figure}[t]
    \begin{center}
    \includegraphics[width=0.85\linewidth]{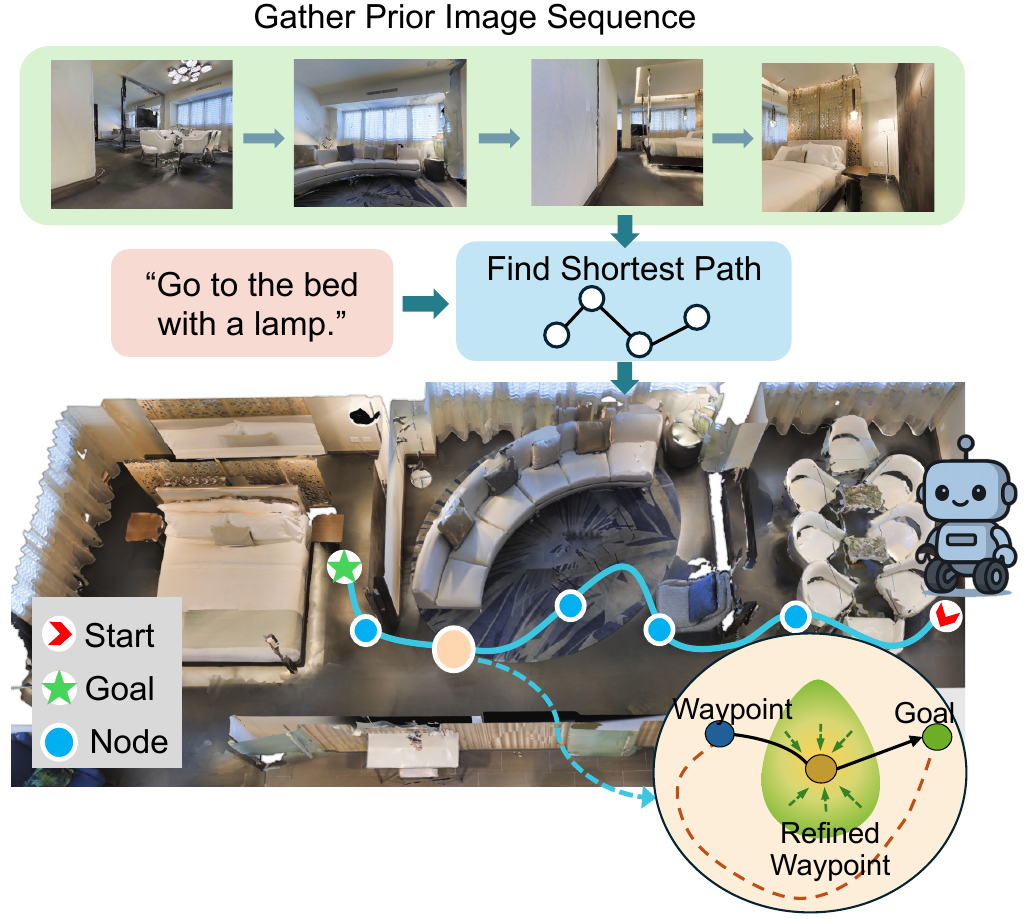}
    \end{center}
    \caption{System workflow. An offline-collected image sequence serves as a topological map. A human instruction specifies the target, from which the shortest path to the goal is planned. An image-goal navigation model produces an initial waypoint, refined by our refinement module for higher-quality guidance.}
    \label{fig:workflow}
\end{figure}

Image-goal navigation using lightweight topological maps
addresses the localization dependency directly. The map
consists solely of geotagged images collected along
traversable paths, making construction, modification, and
extension straightforward. Localization reduces to matching
the current egocentric view against stored reference images,
depending on relative visual similarity rather than absolute
pose accuracy. Unlike object-goal and vision-language navigation, it avoids inefficient mapless exploration and the need for detailed step-by-step instructions. Recent learned waypoint predictors
such as GNM~\cite{gnm} and ViNT~\cite{vint}, trained on
large-scale diverse datasets, can generate ego-centric
navigation targets directly from visual observations,
enabling deployment across different robot platforms in different environments.

However, deploying these models in real-world cluttered
environments exposes two practical limitations. Learned
waypoints are derived purely from image-level features and
carry no explicit geometric guarantees: they can fall on
obstacles, violate traversability constraints, or deviate
from the intended global direction. Meanwhile, translating
these waypoints into safe robot motion requires a reactive
local planner, yet most image-goal navigation systems either
lack such a planner or rely on fixed-parameter controllers
that cannot adapt to confined spaces. 
Rather than discarding the learned prediction in favor of
purely geometric planning, we leverage local depth and
semantic information from the onboard RGB-D sensor to
validate and correct it. 
This motivates a refinement mechanism that
preserves the learned prior while correcting its geometric
deficiencies, paired with a local planner that adapts its
behavior to the surrounding geometry. 
To this end, we introduce EgoNav, a hierarchical navigation system that bridges learned visual guidance with geometry-aware local control. Built on a lightweight topological map, it refines learned waypoints and executes them with an adaptive local planner. The overall workflow is illustrated in Fig.~\ref{fig:workflow}.

Our contribution is at the system level: we integrate
established components through a decision-layer coupling
absent from prior image-goal navigation pipelines.
Specifically:
\begin{enumerate}
\item We present EgoNav, a hierarchical navigation system in
which learned waypoint prediction, geometry-informed
refinement, and adaptive local planning are coupled through
an explicit information flow, where refinement outcomes
directly modulate the local planner.
\item We design a geometry-informed waypoint refinement module
that samples candidate waypoints from traversable regions and
selects reliable targets through multi-criteria energy scoring,
considering geometric safety, global-direction coherence,
and fidelity to the learned prior.
\item We validate the proposed system through extensive experiments
in simulation and on a real humanoid robot, demonstrating
consistent improvements in navigation success rate and path
efficiency over contemporary baselines, with progressive ablations
quantifying each module's contribution.
\end{enumerate}

\section{Related Work}
\label{sec:related_works}

This section reviews the prior work most relevant to EgoNav,
covering visual goal-driven navigation and local path
planning, to contextualize our contributions.

\subsection{Visual Goal-Driven Navigation}
\label{subsec:related_e2e}

Visual navigation research explores various ways to specify
goals, each presenting unique challenges and prompting
distinct methodologies.

\noindent\textbf{Image-Goal Navigation (ImageNav).}
In this task, agents navigate to a location specified by a
target image~\cite{navigating, vint, gnm}. Some methods
construct explicit environment representations online:
ViKiNG~\cite{viking} incrementally builds topological graphs
with heuristic search, while
DeepExplorer~\cite{deepexplore} learns metric-free
topological maps in feature space. GNM~\cite{gnm} achieves broad
cross-platform generalization by training on large-scale
datasets from diverse robot platforms, and ViNT~\cite{vint}
extends it with a Transformer-based architecture for improved
generalization. Unlike these methods, which select subgoal
images at fixed intervals, PlaceNav~\cite{placenav} uses visual
place recognition (VPR) to retrieve the most relevant subgoal
image from a prior map.
However, these pipelines execute the predicted waypoints
as-is: the outputs of~\cite{gnm, vint, placenav} carry no
geometric validation, and no mechanism corrects them against
local traversability. EgoNav adopts the same VPR-based retrieval as
PlaceNav but adds a geometry-informed refinement layer on top
of this pipeline (Section~\ref{sec:waypoint}).
\vspace{0.5em}

\noindent\textbf{Object-Goal Navigation (ObjectNav).}
ObjectNav specifies the goal as an object category, requiring
exploration and semantic
understanding~\cite{habitat-web, aligning}. Modular approaches
guide exploration with learned potential functions~\cite{poni},
online semantic maps~\cite{sem-object}, or vision-language
frontier values~\cite{vlfm}. Without a prior map, however, the
task often degenerates into inefficient exploration, especially
over long distances.
\vspace{0.5em}

\noindent\textbf{Vision-Language Navigation (VLN).}
VLN agents follow natural language instructions, demanding
cross-modal spatial grounding~\cite{vlnsurvey}. Recent LLM-based
methods improve reasoning, memory, and safety~\cite{vln:navgpt,
mcgpt, safevln}, or adopt modular and hierarchical designs for
long-horizon tasks~\cite{msnav, nava3}. Yet the dependency on
detailed step-by-step instructions remains impractical to
satisfy at scale.

\subsection{Local Path Planning}
\label{subsec:related_local_planning}

Local path planning generates feasible short-term trajectories
toward immediate goals while reacting to sensor data. Existing
strategies span a broad spectrum: reactive methods~\cite{lp:simple}
guarantee completeness through boundary following but often
produce suboptimal paths, sampling-based approaches such as
Falco~\cite{falco} achieve high speed by evaluating pre-computed
motion primitives, optimization-based techniques including
egoTEB~\cite{egoteb} refine trajectories toward local optimality
at higher computational cost, and path deformation
methods~\cite{lp:reshaping} adapt a reference path smoothly
around obstacles. In settings where accurate 3D maps and precise
localization are available, trajectory optimization methods such
as Ego-Planner~\cite{planner:ego-planner} and
SVSDF~\cite{planner:svsdf} can generate globally informed
collision-free paths, but these prerequisites are hard to obtain in real practice. 
EgoNav builds upon Falco for its real-time efficiency and
operates without a metric map; unlike prior systems that run
the local planner with fixed parameters decoupled from
upstream prediction, it modulates planner behavior based on
the refinement outcomes (Section~\ref{sec:local_planner}).
\begin{figure*}[ht]
    \centering
    \includegraphics[width=0.95\linewidth]{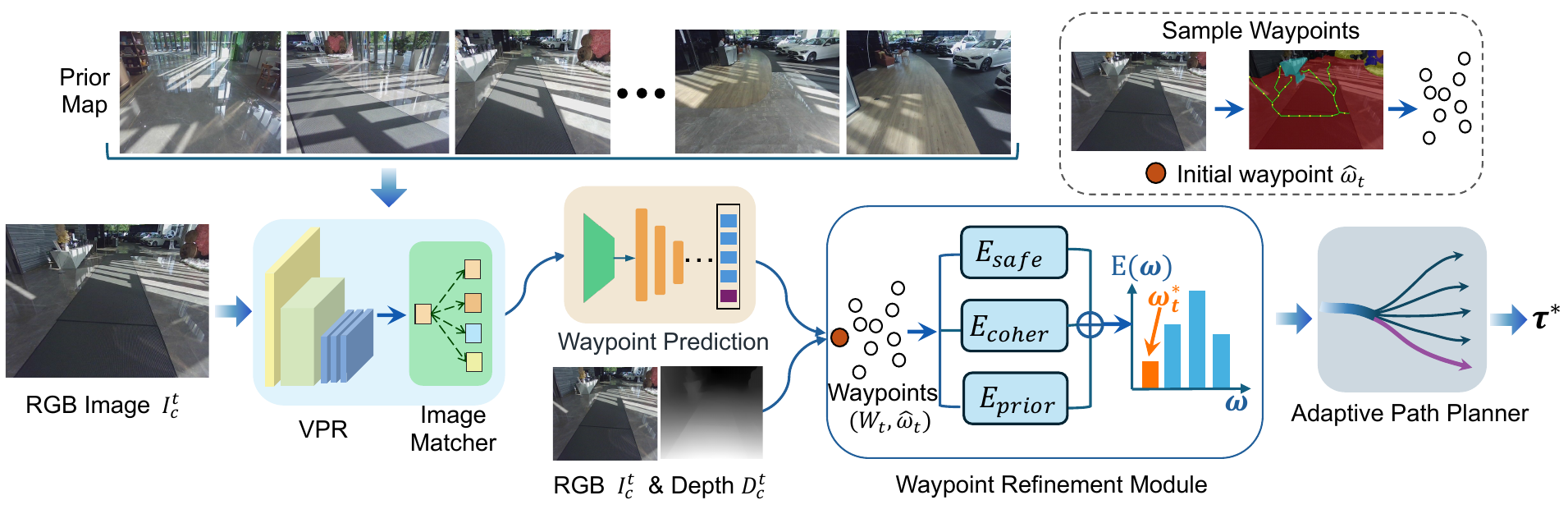}
    \caption{
    The EgoNav algorithm pipeline. At timestep $t$, the system receives the RGB image $I_c^t$ and depth image $D_c^t$ as inputs. The VPR module encodes $I_c^t$ and matches it against a pre-collected image sequence to retrieve a subgoal. This subgoal, together with the current image, is fed into GNM to produce an initial waypoint $\hat{w}_t$. The candidate set $\mathcal{W}_t$, which includes $\hat{w}_t$, is then evaluated and refined by the waypoint refinement module through multi-criteria scoring. The corrected waypoint $w_t^*$ is passed to an adaptive path planner to generate the optimal trajectory $\tau_t^*$.
    }
    \label{fig:pipeline}
\end{figure*}

\section{Methodology}
\label{sec:methodology}

\subsection{Problem Formulation}
\label{sec:overview}

We consider a robot equipped with an RGB-D sensor navigating
in an indoor environment using a pre-collected topological map
$\mathcal{M} = \{(I_n, p_n)\}_{n=1}^{N}$, where each node
stores a reference image $I_n$ and its pose
$p_n = (x, y, \theta)_n$. Edges connect temporally adjacent nodes recorded along the traversed collection trajectory, which guarantees that connected nodes are mutually reachable.
When a location is revisited, coincident nodes are merged into one node that retains both views, 
joining separate segments into a connected graph rather than a single path.
Given a target image $I_g \in \{I_n\}$, the robot must reach
the corresponding location by generating real-time, collision-free
velocity commands from its RGB observation $I^t_c$ and depth
image $D^t_c$ at each timestep $t$.

EgoNav adopts a hierarchical strategy that decomposes this task
into three stages (Fig.~\ref{fig:pipeline}).
A subgoal retrieval module localizes the robot within the
topological map via VPR and selects
an appropriate subgoal along the planned path
(\S\ref{sec:global}).
A pre-trained waypoint prediction network then produces
an initial waypoint toward the subgoal, and a multi-criteria
refinement module evaluates this prediction alongside
geometrically grounded alternatives to enforce safety and
directional consistency (\S\ref{sec:waypoint}).
An adaptive local path planner finally generates a collision-free
trajectory whose parameters are calibrated to both
environmental geometry and upstream correction outcomes
(\S\ref{sec:local_planner}).

\subsection{Map-Guided Waypoint Prediction}
\label{sec:global}

\paragraph{Active Subgoal Retrieval}
GNM~\cite{gnm} is a pre-trained network that predicts an
ego-centric waypoint from a current observation paired with a
subgoal image, encoding useful priors such as center-biased
and smooth motion. In its original formulation, 
however, GNM retrieves subgoal images sequentially 
from the recorded trajectory, implicitly assuming the robot tracks the
recorded route at a uniform pace, and misaligns once the
robot deviates or changes speed.
We adopt active subgoal retrieval similar to
PlaceNav~\cite{placenav}. At startup, the robot localizes
within $\mathcal{M}$ using sequence-based VPR~\cite{iloc}
with MixVPR~\cite{mixvpr} as the feature backbone, and
computes the shortest topological path
$\Gamma = [M_1, \ldots, M_{\text{target}}]$ via Dijkstra's
algorithm. Since nodes are equally spaced at fixed interval,
all edges share equal weight, so Dijkstra's shortest path
yields the shortest traversable route. During navigation,
$I^t_c$ is continuously matched against nodes within a
sliding window along $\Gamma$, producing a matched index
$m_t$; the node a fixed number of steps ahead is selected as
the subgoal $I^t_s$ and paired with $I^t_c$ as GNM input,
yielding an ego-centric waypoint
$\hat{w}_t = (x, y, \Delta\theta)_t$.
This VPR-based retrieval keeps the subgoal image aligned with the
actual position of the robot.

\paragraph{Localization Filtering and Relocalization}
VPR matching is susceptible to localization errors,
particularly in environments with repetitive textures or
lighting changes. The most disruptive case is backward drift,
where perceptual aliasing causes $m^{\text{raw}}_t$ to
regress to a previously visited node. Even a shift of one or
two nodes backward pulls the subgoal toward an already-passed
location and may direct the robot to retrace its path.
We suppress this by recording the previous matched index
$m_{t-1}$ and enforcing $m_t = \max(m^{\text{raw}}_t,\;
m_{t-1})$, ensuring that the index never reverts to a node
before the most recently visited one.

When the localization error is large, for instance a spurious
match that jumps many nodes forward due to a visually similar
but distant location, clamping alone is insufficient.
We detect such cases by checking whether $m_t$ advances more
than $j_{\max}$ nodes beyond $m_{t-1}$ in a single cycle.
Since the robot advances only a few nodes per cycle, $j_{\max}$ is set to admit this range plus minor matching fluctuation while rejecting implausible jumps. When this condition is violated, the robot executes a backtracking procedure by retracing its recent path using
buffered velocity commands recorded during navigation. After
backtracking, it performs initial localization by capturing
a short image sequence and matching it against the full map
$\mathcal{M}$ using sequence-based VPR, after which $\Gamma$
is replanned from the recovered node.

\bigskip
While this filtering stabilizes the subgoal input to GNM, the
predicted waypoint $\hat{w}_t$ can still be unreliable.
Since GNM is trained on image-level features without
explicit geometric reasoning, it may produce waypoints that
fall on or near obstacles in cluttered scenes, deviate from
the intended global direction under visual ambiguity, or fail
to generalize to previously unseen environments. These failure
modes cannot be resolved by improving the subgoal input alone;
 they necessitate a correction mechanism that incorporates local geometric information derived from the depth and segmentation observations available at each timestep.

\subsection{Geometry-Informed Waypoint Refinement}
\label{sec:waypoint}
While learned waypoint predictors may produce geometrically
unsafe or directionally inconsistent targets, they also encode
useful navigation priors such as center-biased and smooth
motion preferences. Therefore, refinement should correct
geometric inconsistencies while preserving these priors.
The refinement module converts $\hat{w}_t$ into a corrected
waypoint $w_t^*$ that is geometrically safe, aligned with the
global path, and close to the original prediction.

\paragraph{Traversability-Guided Candidate Sampling}
We apply the RGB-D semantic segmentation model
DFormerV2~\cite{dformerv2} to $(I^t_c, D^t_c)$, producing
a binary traversability map $M_{\text{trav}}$ in the local
frame. As a safeguard against segmentation errors, a
cross-modal consistency check compares traversability labels
with raw depth readings: if more than a fraction $\gamma$ of
pixels labeled traversable have depth values below
$d_{\text{safe}}$, the segmentation is deemed unreliable and
$\hat{w}_t$ is returned directly without refinement.
When the check passes, we extract the morphological skeleton
of $M_{\text{trav}}$ through iterative thinning, which
progressively erodes boundary pixels while preserving
topological connectivity, yielding a one-pixel-wide medial
axis of the free space. Candidate waypoints,  each represented as a 2-D position $(x,y)$ in the local frame, are sampled along this skeleton at intervals of
0.25\,m. Sampling along the medial axis biases candidates toward the
center of free space, which empirically improves corridor traversal
and doorway passage compared with uniform grid sampling.
The GNM prediction $\hat{w}_t$ is included in the resulting candidate set
$\mathcal{W}_t$, so that if the learned prediction is already
adequate, it is selected naturally through the scoring
process.

\paragraph{Waypoint Scoring and Selection}
Each candidate waypoint $w \in W_t$ is evaluated using a weighted
energy function that corrects the above failure modes while
preserving the useful prior encoded in the learned waypoint predictor:
\begin{equation}
    E(w) = \lambda_s \, E_{\text{safe}}(w)
         + \lambda_c \, E_{\text{coher}}(w)
         + \lambda_p \, E_{\text{prior}}(w).
    \label{eq:total_cost}
\end{equation}

To avoid obstacle-intersecting predictions, 
the safety term penalizes candidates close to obstacles:
\begin{equation}
    E_{\text{safe}}(w) = \sigma\bigl(
        d_{\text{safe}} - d(w);\, k\bigr),
    \label{eq:safety}
\end{equation}
where $d(w)$ is the distance from $w$ to the nearest obstacle
boundary. To reduce sensitivity to per-pixel segmentation
noise, $d(w)$ is computed as a weighted average over a local
window around $w$, where each pixel is weighted by its
traversability confidence score from DFormerV2.
The parameter $d_{\text{safe}}$ is the clearance threshold
derived from the footprint radius of the robot, and
$\sigma(x; k) = 1/(1+e^{-kx})$ is a logistic function with
steepness $k$. Candidates farther than $d_{\text{safe}}$ from
obstacles receive near-zero cost, while those within the
safety margin are penalized sharply.

To reduce directional drift at turning points, 
the coherence term encourages alignment with the global navigation direction:
We define the guidance vector $\mathbf{v}_g$ as the unit vector from the current pose toward
the subgoal node along $\Gamma$:
\begin{equation}
    E_{\text{coher}}(w) = \frac{1}{2}
        \!\left(1 - \frac{\mathbf{v}_w \cdot \mathbf{v}_g}
        {\|\mathbf{v}_w\| \, \|\mathbf{v}_g\|}\right),
    \label{eq:coherence}
\end{equation}
where $\mathbf{v}_w$ is the vector from the current pose to
candidate $w$, a value of zero indicates perfect alignment
with the global direction. Here both poses are pre-stored: the current pose is that of the currently VPR-matched node, and the subgoal pose that of the subgoal node along $\Gamma$; candidate waypoints are transformed into the same frame via this matched-node pose,
so the two vectors are directly comparable.

To preserve useful navigation priors, 
the prior term penalizes excessive deviation from the predicted waypoint:
\begin{equation}
    E_{\text{prior}}(w) = \sigma\bigl(
        \|w - \hat{w}_t\| - d_{\text{prior}};\, k\bigr),
    \label{eq:prior}
\end{equation}
where $d_{\text{prior}} = 2 \, d_{\text{safe}}$.
Since $\hat{w}_t \in \mathcal{W}_t$ always receives
$E_{\text{prior}} = 0$, the learned prediction is granted an inherent advantage that is overridden only when geometric
evidence strongly favors an alternative candidate.

The above energy terms are normalized to the range $[0,1]$,
enabling consistent weighting through the coefficients $\lambda$.
The corrected waypoint is then selected as
$w_t^* = \arg\min_{w \in \mathcal{W}_t} E(w).$
Along with the selected waypoint, two auxiliary quantities
computed during scoring are retained: the correction magnitude
$\delta_t = \|w_t^* - \hat{w}_t\|$, which measures the deviation
from the original prediction, and the waypoint clearance
$d(w^*_t)$, which indicates the distance of the corrected
waypoint to the nearest obstacle.

\subsection{Collision-Free Path Generation}
\label{sec:local_planner}

The corrected waypoint $w^*_t$ defines a local navigation
goal but does not itself constitute a directly executable trajectory. To bridge this gap, the local path planner translates $w^*_t$ into a collision-free velocity
command using the depth-derived point cloud $P^t_c$.


\paragraph{Trajectory Library Evaluation}
We adopt Falco~\cite{falco} as the base planner for its
real-time efficiency. Falco evaluates a pre-computed library
of kinodynamically feasible trajectories organized into
discrete path groups, each representing a distinct motion
intention; a spatial lookup structure enables fast collision
checking against obstacle points within the planning scope.
At runtime, each group is scored by aggregating the
goal-reaching likelihood of its collision-free trajectory
endpoints, where endpoints closer to and oriented toward the
refined waypoint $w^*_t$ receive higher likelihood, and the
highest-scoring group determines the executed trajectory. If
no collision-free path exists within the current scope, the
planner progressively retracts the search horizon until a
feasible alternative is found.

\paragraph{Adaptive Planner Modulation}
The standard Falco planner uses fixed parameters, which forces a global
compromise between agility in open spaces and caution in confined areas. We
address this by dynamically modulating a small set of planner parameters using
two byproducts of the upstream correction module: the waypoint clearance
$d(w^*_t)$ and the correction magnitude $\delta_t$.

Each is converted into a normalized factor $g\in[0,1]$ indicating how strongly
to modulate:
\begin{equation}
g_c=\mathrm{clamp}\!\Big(1-\tfrac{d(w^*_t)}{d_{\text{safe}}},0,1\Big),\qquad
g_\delta=\mathrm{clamp}\!\Big(\tfrac{\delta_t}{\delta_{\text{ref}}},0,1\Big),
\end{equation}
where $d_{\text{safe}}$ and $\delta_{\text{ref}}$ are reference thresholds;
$g_c$ grows as the waypoint approaches obstacles and $g_\delta$ as the
correction grows, both vanishing in open, well-predicted conditions. Each
affected parameter is then linearly interpolated between its default and its
bound:
\begin{equation}
\begin{aligned}
S &= S_0 - g_c\,(S_0-S_{\min}), &N &= N_0 + g_c\,(N_{\max}-N_0),\\
R &= R_0 - g_\delta\,(R_0-R_{\min}), &V &= V_0 - \max(g_c,g_\delta)\,(V_0-V_{\min}),
\end{aligned}
\end{equation}
where $S$, $N$, $R$, and $V$ are the trajectory scale, collision tolerance
(obstacle points admitted per path), planning horizon, 
and maximum velocity, respectively. 
A small clearance thus reduces the planning scope and raises
the collision tolerance, admitting compact trajectories that
pass closer to boundaries and traverse narrow passages pruned
under default settings; a large correction shortens the
planning horizon for finer, step-by-step planning. 
Driven by $\max(g_c,g_\delta)$, the velocity drops whenever space is confined or correction is large, pairing any tolerance relaxation with a slowdown.

This calibrates the planner not only on perceived geometry but on how
much the upstream module had to intervene. The optimal trajectory $\tau^*_t$ is
selected from the highest-scoring motion group, and the corresponding velocity
command $(v_t,\omega_t)$ is sent to the low-level controller for execution.

\begin{figure}[ht]
    \centering
    \includegraphics[width=0.85\linewidth]{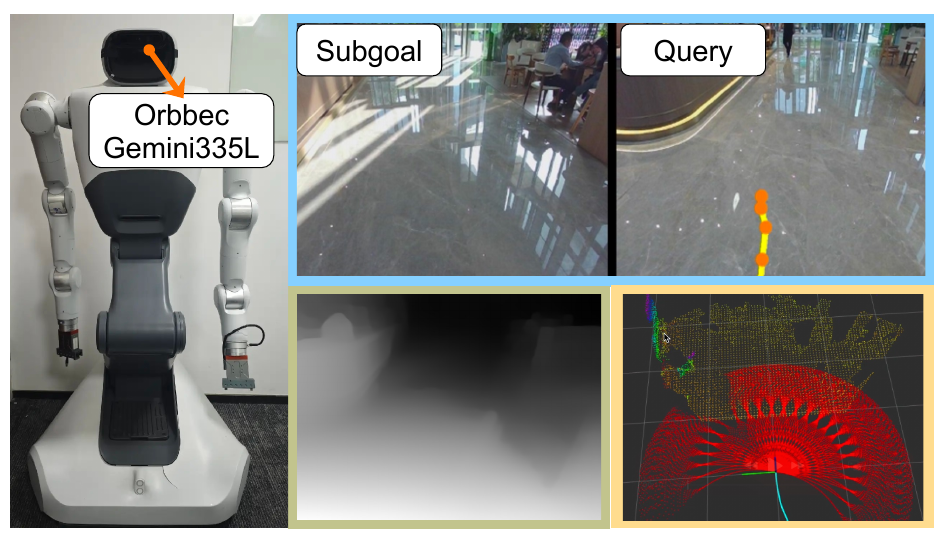}
    \caption{Physical platform and interface. Left: the humanoid robot used for real-world validation. Right: the current query image $I_c^t$, VPR-retrieved subgoal $I_s^t$, and GNM-predicted waypoint (blue); the depth map (green); and collision-free paths from our local planner (yellow).}
    \label{fig:human_robot}
\end{figure}

\section{Experiments}
\label{sec:experiments}

We evaluate EgoNav in simulation and real world,
comparing against contemporary baselines and conducting
ablation studies to analyze the contribution of each component.

\subsection{Experimental Setup}
\label{sec:setup}

\paragraph{Platforms}

Simulated experiments use Habitat-sim~\cite{habitat} with Matterport3D (MP3D)~\cite{mp3d} and a LoCoBot with an RGB-D sensor (90$^\circ$ FOV). Real-world experiments use a humanoid robot on an omnidirectional chassis with an Orbbec Gemini335L RGB-D camera (Fig.~\ref{fig:human_robot}), running inference on a Jetson Orin NX and control on an Intel NUC.

\paragraph{Metrics}

We report Success Rate (SR), the fraction of episodes reaching within 1\,m of the target without collision, and Success weighted by inverse Path Length (SPL)~\cite{habitat}, the shortest-to-actual path-length ratio over successful episodes, jointly capturing success and efficiency. The shortest path is the Dijkstra path on the topological graph, used identically in simulation and the real world.

\paragraph{Datasets}
In simulation, we use 12 MP3D scenes with varied layouts (open playrooms, compact bedrooms, L-shaped corridors, multi-room apartments with successive doorways, etc.). Start and goal positions are randomly sampled along traversable paths, with RGB-D images and poses recorded at 0.25\,m intervals. Geodesic distance defines three difficulty levels: short ($<$10\,m), medium (10--20\,m), and long ($>$20\,m), 30 episodes per method per level. Real-world evaluation covers 7 scenes over two floors of an office building: a lobby, open-plan workspace, corridors with 90$^\circ$ turns, and offices of varying furniture density.

\paragraph{Baselines}
We compare against methods from two navigation paradigms.
Among image-goal methods that use a prior topological map,
GNM~\cite{gnm}, ViNT~\cite{vint}, NoMaD~\cite{nomad},
PlaceNav~\cite{placenav}, and DeepExplore~\cite{deepexplore}
predict waypoints or actions from visual inputs.
As an exploration-based baseline, VLFM~\cite{vlfm} navigates
without a prior map through frontier-based exploration guided
by a vision-language model, using depth for obstacle avoidance.
To isolate the contribution of our proposed modules, we
additionally evaluate PlaceNav+Falco: PlaceNav augmented with
the same Falco local path planner used in EgoNav but with
default fixed parameters. This baseline uses the same
perception inputs as EgoNav (RGB-D) but does not include
waypoint refinement, adaptive planner modulation, or
relocalization, ensuring that any performance gap is
attributable to these modules.

\begin{table}[t]
\centering
\caption{Parameter settings of EgoNav.}
\label{tab:params}
\setlength{\tabcolsep}{4pt}
\begin{tabular}{lll}
\hline
Parameter & Value & Derivation \\
\hline
$d_{\text{safe}}$ & 0.8\,m & Robot footprint + margin \\
$d_{\text{prior}}$ & 1.6\,m & $2 \, d_{\text{safe}}$ \\
$j_{\max}$ & 4 & Max plausible advance per cycle \\
\hline
$\lambda_s$ & 0.5 & Grid search \\
$\lambda_c$ & 1.5 & Grid search \\
$\lambda_p$ & 1.0 & Grid search \\
$k$ & 3 & Grid search \\
$\gamma$ & 0.2 & Grid search \\
\hline
\end{tabular}
\end{table}

\paragraph{Implementation Details}
For image-goal methods with prior maps, the target is
specified directly as a goal image selected from the
topological map. For VLFM, the target is specified as an
object name corresponding to a distinctive landmark visible
at the goal location, chosen to avoid ambiguity with other
instances of the same object category in the environment.
All learning-based methods use their publicly released
pre-trained weights without fine-tuning on any evaluation
environment.

The parameters of refinement and planning modules are
fixed across all simulation and real-world
experiments. The safety threshold $d_{\text{safe}}$ is set to 
the robot's footprint radius plus a fixed margin for both 
the simulated LoCoBot and the physical humanoid robot. 
Table~\ref{tab:params} summarizes their values
and derivations. The upper group of parameters is derived from physical
quantities or map resolution and requires no tuning. 
The lower group contains the hyperparameters determined
empirically on a held-out set of five simulation scenes
disjoint from the twelve evaluation scenes, by sweeping
candidate values for each parameter and selecting the
combination that maximized SR on medium-distance episodes.

\begin{figure*}[t]
    \centering
    \includegraphics[width=0.9\linewidth]{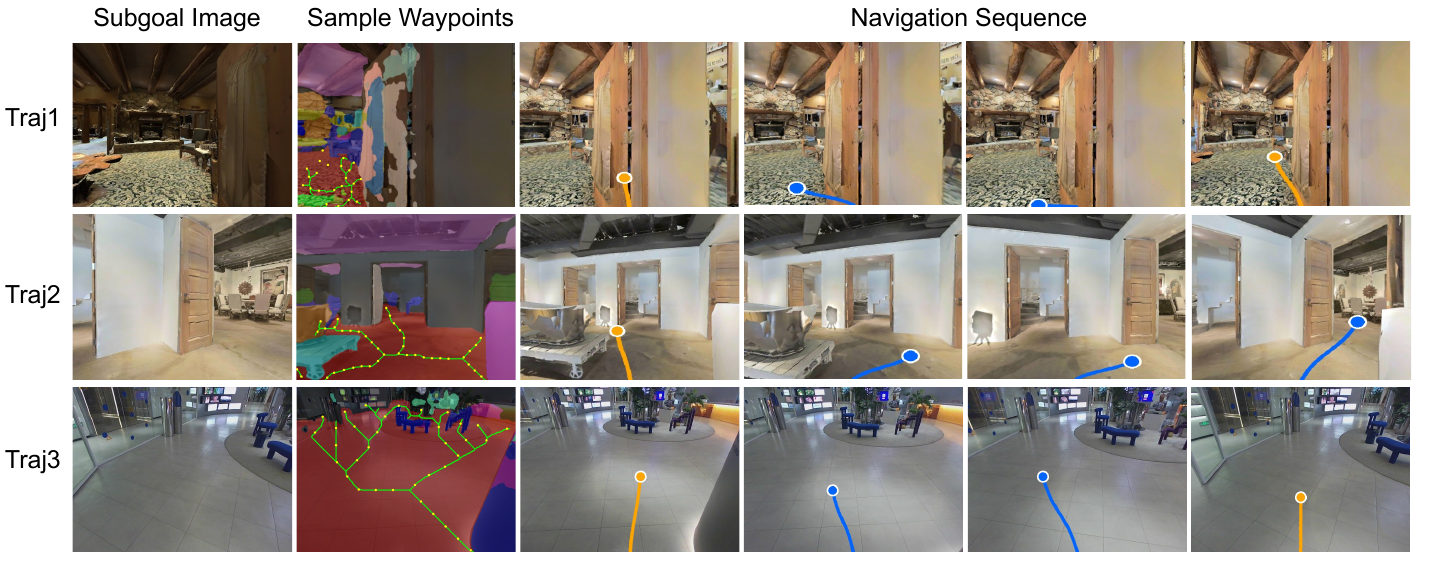}
    \caption{Waypoint refinement in action. Each row shows a
    navigation trajectory. The first column shows the subgoal
    image; the second shows semantic segmentation with skeleton
    extraction and candidate sampling. 
    The following four columns show consecutive timesteps.
    Orange dots indicate the initial waypoint was selected;
    blue dots indicate a skeleton-sampled candidate was
    selected instead. The
transition from orange to blue across frames reflects the
robot turning to follow the refined direction.}
    \label{fig:pwc}
\end{figure*}

\subsection{Main Results}
\label{sec:results}

\begin{table}[t]
\centering
\caption{Comparison with baselines in Habitat-sim (MP3D),
30 episodes per method per difficulty level.}
\label{tab:sim}
\setlength{\tabcolsep}{3pt}
\begin{tabular}{l|cc|cc|cc}
\hline
\multirow{2}{*}{Method}
  & \multicolumn{2}{c|}{$<$10\,m}
  & \multicolumn{2}{c|}{10--20\,m}
  & \multicolumn{2}{c}{$>$20\,m} \\
 & SR & SPL & SR & SPL & SR & SPL \\
\hline
GNM~\cite{gnm}             & 63.3 & 61.0 & 36.7 & 34.5 & 26.7 & 24.2 \\
ViNT~\cite{vint}            & 60.0 & 57.8 & 33.3 & 31.0 & 20.0 & 17.6 \\
PlaceNav~\cite{placenav}    & 70.0 & 67.4 & 46.7 & 44.0 & 36.7 & 33.1 \\
NoMaD~\cite{nomad}          & 63.3 & 60.5 & 33.3 & 30.2 & 13.3 & 11.5 \\
DeepExplore~\cite{deepexplore} & 23.3 & 21.6 & 10.0 & 9.2 & 0.0 & 0.0 \\
\hline
VLFM~\cite{vlfm}            & 56.7 & 40.1 & 43.3 & 24.6 & 26.7 & 9.8 \\
\hline
PlaceNav+Falco             & 80.0 & 74.3 & 56.7 & 50.2 & 46.7 & 41.0 \\
EgoNav (Ours)               & 96.7 & 89.1 & 76.7 & 64.9 & 60.0 & 50.8 \\
\hline
\end{tabular}
\end{table}

\paragraph{Simulation}
Table~\ref{tab:sim} presents the simulation results. Among
image-goal methods without a local path planner, PlaceNav performs
best due to VPR-based subgoal selection, while the other methods
degrade rapidly at longer distances where collisions become
the dominant failure mode. DeepExplore, which relies on
discrete actions without depth sensing, fails entirely beyond
20\,m. 
A consistent pattern across map-guided methods is that SPL
tracks SR relatively closely at short range, since successful
episodes follow the pre-planned topological path and produce
trajectories near the shortest route. The gap between SR and
SPL widens at longer distances as minor detours accumulate
over more waypoints.
VLFM presents a sharp contrast, with its SPL falling far
below SR at every distance level, reflecting the path
inefficiency inherent in frontier-based exploration without
a prior map.

PlaceNav+Falco, the sensor-fair baseline, confirms that
depth-based local planning alone provides a substantial
improvement over methods that execute learned waypoints
directly. EgoNav further outperforms PlaceNav+Falco at every
distance level, with the advantage remaining consistent
from short to long range. This indicates that waypoint
refinement and correction-informed modulation address failure modes
that the local path planner cannot resolve on its own, primarily
directional errors at turning points and overly conservative
collision checking in narrow passages.

\begin{table}[t]
\centering
\caption{Comparison with baselines in real-world environments,
30 episodes per method per difficulty level.}
\label{tab:real}
\setlength{\tabcolsep}{3pt}
\begin{tabular}{l|cc|cc|cc}
\hline
\multirow{2}{*}{Method}
  & \multicolumn{2}{c|}{$<$10\,m}
  & \multicolumn{2}{c|}{10--20\,m}
  & \multicolumn{2}{c}{$>$20\,m} \\
 & SR & SPL & SR & SPL & SR & SPL \\
\hline
GNM~\cite{gnm}             & 50.0 & 47.6 & 26.7 & 24.5 & 16.7 & 14.3 \\
ViNT~\cite{vint}            & 46.7 & 44.0 & 23.3 & 21.2 & 16.7 & 11.0 \\
PlaceNav~\cite{placenav}    & 50.0 & 47.5 & 36.7 & 32.8 & 26.7 & 23.4 \\
NoMaD~\cite{nomad}          & 50.0 & 45.5 & 26.7 & 24.1 & 16.7 & 13.8 \\
DeepExplore~\cite{deepexplore} & 16.7 & 15.3 & 6.7 & 6.0 & 0.0 & 0.0 \\
\hline
VLFM~\cite{vlfm}            & 53.3 & 44.8 & 26.7 & 13.6 & 16.7 & 5.9 \\
\hline
PlaceNav+Falco             & 60.0 & 57.2 & 50.0 & 44.8 & 33.3 & 28.2 \\
EgoNav (Ours)               & 73.3 & 65.6 & 66.7 & 58.1 & 46.7 & 34.0 \\
\hline
\end{tabular}
\end{table}

\paragraph{Real-World}
Table~\ref{tab:real} reports performance on the physical
humanoid robot. The ranking among methods is consistent with
simulation, and the gap between PlaceNav+Falco and
EgoNav persists across all distances. All methods show a
moderate drop compared to simulation, primarily due to
lighting differences between map collection and navigation
time that degrade VPR accuracy. This effect accumulates over
longer paths, making the performance decline steeper at long
range. 
EgoNav's localization filtering and relocalization
mechanism mitigates this issue to some extent by detecting
and recovering from severe matching failures. 
Specifically, when the robot deviates from the 
mapped route due to VPR mismatch or obstacle-induced detours, 
backtracking retraces the recent path and reattempts localization. 
Since the replay is open-loop, however, 
accumulated error limits its effectiveness to short detours of a few meters.

\begin{figure}[t]
    \centering
    \includegraphics[width=0.85\linewidth]{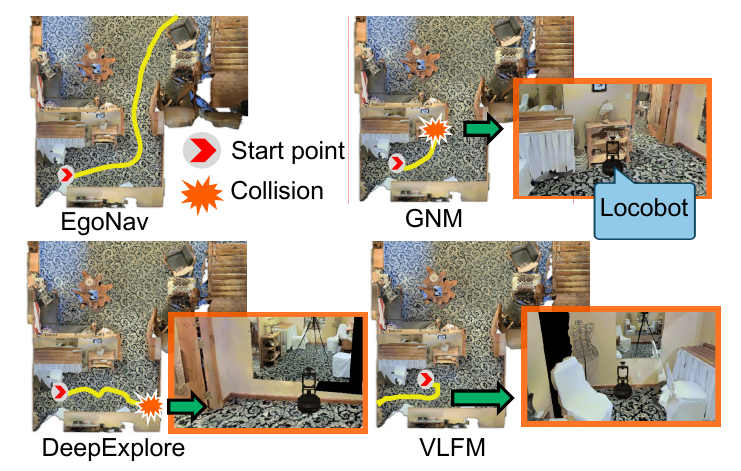}
    \caption{Qualitative comparison of navigation paths in
    simulation. The task requires navigating out of a narrow
    room through a doorway. EgoNav finds a collision-free path,
    while GNM and DeepExplore collide with obstacles (orange
    starburst) shortly after starting. VLFM avoids collisions
    but becomes trapped in a corner.}
    \label{fig:traj}
\end{figure}

Fig.~\ref{fig:pwc} shows the refinement module on three trajectories (Traj1--2 simulated, Traj3 real). In Traj1, a GNM waypoint on a doorframe is refined to the opening center. In Traj2 and Traj3, the GNM waypoint points straight ahead at a sharp turn while the subgoal lies to the side; the coherence term steers selection back onto the global path, preventing overshoot.
Fig.~\ref{fig:traj} compares full paths through a narrow doorway: EgoNav stays collision-free, GNM and DeepExplore collide shortly after departure, and VLFM avoids collisions but gets trapped in a corner.

\subsection{Ablation Studies}
\label{sec:ablation}

All ablation experiments are conducted in simulation on the
medium-distance (10--20\,m) episodes with 30 runs.

Table~\ref{tab:progressive} shows the incremental impact of
each component. Each row adds one module on top of the previous one.
Overall navigation success improves consistently as additional
modules are introduced, while SPL reflects the trade-off between
success recovery and path efficiency, particularly when
relocalization is triggered.
VPR-based subgoal retrieval replaces naive sequential traversal
with position-aware selection.
The local path planner converts waypoint commands into
collision-free trajectories.
Waypoint refinement improves alignment with the global path,
especially at turning points where GNM tends to point straight ahead.
Adaptive planner modulation improves path efficiency with
success rate unchanged: relaxing collision sensitivity and
refining search granularity near low-clearance regions yields
shorter, smoother trajectories through narrow passages.
Finally, localization filtering and relocalization stabilize
subgoal tracking and recover from large localization errors,
leading to the highest success rate.

\begin{table}[t]
\centering
\caption{Progressive ablation of EgoNav components
on the 10--20 m navigation task (30 runs).}
\label{tab:progressive}
\begin{tabular}{l|cc}
\hline
Configuration & SR ($\uparrow$) & SPL ($\uparrow$) \\
\hline
GNM                            & 36.7 & 34.5 \\
+ VPR retrieval (PlaceNav)               & 46.7 & 44.0 \\
+ Local path planner (PlaceNav + Falco)           & 56.7 & 50.2 \\
+ Waypoint refinement          & 70.0 & 61.2 \\
+ Adaptive Planner Modulation  & 70.0 & 62.6 \\
+ Localization filtering \& relocalization (Full) & 76.7 & 64.9 \\
\hline
\end{tabular}
\end{table}

\begin{table}[t]
\centering
\caption{Ablation of individual refinement criteria
on the 10--20 m navigation task (30 runs).}
\label{tab:loss_ablation}
\begin{tabular}{l|cc}
\hline
Configuration & SR ($\uparrow$) & SPL ($\uparrow$) \\
\hline
Full refinement                         & 76.7 & 62.4 \\
w/o $E_{\text{coher}}$                  & 56.7 & 48.0 \\
w/o $E_{\text{prior}}$                  & 63.3 & 54.5 \\
w/o $E_{\text{safe}}$                   & 76.7 & 59.8 \\
w/o refinement entirely                 & 53.3 & 47.9 \\
\hline
\end{tabular}
\end{table}

Table~\ref{tab:loss_ablation} isolates the contribution of each scoring term, evaluated in a separate run from Table IV (hence minor numerical differences). Removing $E_{\text{coher}}$ causes the largest SR drop, confirming directional coherence as the most critical criterion. Without it, the agent overshoots sharp turns and drifts beyond the coverage of the topological map, leading to irrecoverable localization failures. Removing $E_{\text{prior}}$ also noticeably degrades SR: without this anchor to the waypoint prediction, the module selects geometrically safe but positionally unnatural candidates, often hugging walls rather than remaining centered in the traversable area, reducing path quality and occasionally causing jerky planner behavior. Removing $E_{\text{safe}}$ leaves SR unchanged but lowers SPL, as waypoints tend to fall closer to obstacle boundaries, forcing the downstream planner to search for feasible paths in tighter spaces and invoke retraction more frequently. While a viable route can still be found and SR is preserved, the resulting trajectories are less efficient. The weight magnitudes in Table~\ref{tab:params} mirror this ordering: $\lambda_c$ is largest as directional errors force costly global relocalization, while $\lambda_s$ is smallest since safety violations can still be absorbed by the downstream planner's collision checking.
\begin{table}[t]
\centering
\caption{Average latency of core components on the
humanoid platform.}
\label{tab:latency}
\begin{tabular}{l|cccc}
\hline
Component & Preprocess & Waypoint Prediction & Refinement & Planner \\
\hline
Time (ms) & 100 & 20 & 40 & 2.5 \\
\hline
\end{tabular}
\end{table}

Table~\ref{tab:latency} reports the average per-cycle latency
on the humanoid platform. Preprocessing, which includes
semantic segmentation and candidate waypoint sampling, dominates the computation. The total latency of approximately 160\,ms
supports a 5\,Hz control loop, which is sufficient for reactive indoor navigation at walking speed.

\begin{figure}[t]
\centering
\includegraphics[width=\linewidth]{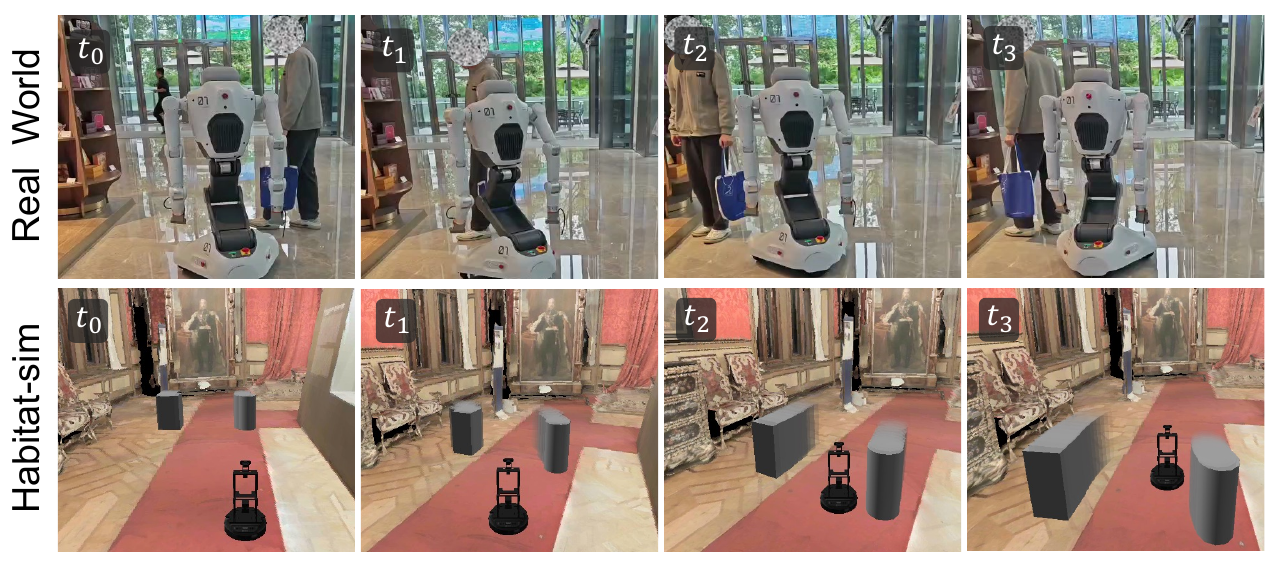}
\caption{Dynamic-obstacle avoidance (top: real; bottom: sim). In simulation, faded copies mark obstacles' past positions.}
\label{fig:dynamic}
\end{figure}

\subsection{Robustness to Dynamic Obstacles}
\label{sec:dynamic}

The preceding experiments are conducted in static scenes. To assess how EgoNav behaves when moving obstacles appear, we place a varying number of dynamic obstacles at random along the robot's path, in both simulation and the real world. Simulation reuses the 30 medium-difficulty (10--20\,m) episodes; the real world uses 10 newly collected episodes over 5 dynamic scenes. We compare against PlaceNav+Falco (Table~VII): since the refinement module only adjusts the waypoint and planner parameters without touching Falco's collision avoidance, it leaves reactive motion handling intact, and EgoNav inherits the same avoidance capability while retaining its static-scene margin (Fig.~\ref{fig:dynamic}).

This avoidance is reactive, inherited from Falco. It suffices for low-density motion, but fast obstacles on random trajectories are hard to evade; moreover, evasive detours can steer the robot off the planned path and trigger relocalization, lowering success. Handling such cases requires velocity-level reasoning and trajectory prediction, left to future work.
\section{Discussion}
\label{sec:conclusion}

\begin{table}[t]
\centering
\caption{Effect of dynamic obstacles (10--20\,m). Sim static rows are taken from the 10--20 m results in Table II. Sim: 30 episodes; real: 10 newly collected episodes over 5 scenes. }
\label{tab:dynamic}
\begin{tabular}{llcccc}
\toprule
& & \multicolumn{2}{c}{PlaceNav+Falco} & \multicolumn{2}{c}{EgoNav (Ours)}\\
\cmidrule(lr){3-4}\cmidrule(lr){5-6}
& Setting & SR & SPL & SR & SPL\\
\midrule
\multirow{2}{*}{Sim}  & Static  & 56.7 & 50.2 & 76.7 & 64.9\\
                      & Dynamic & 40.0 & 32.8 & 63.3 & 50.4\\
\midrule
\multirow{2}{*}{Real} & Static  & 50.0 & 43.1 & 70.0 & 57.3\\
                      & Dynamic & 30.0 & 20.8 & 50.0 & 36.5\\
\bottomrule
\end{tabular}
\end{table}

We present EgoNav, a hierarchical system for image-goal
navigation that bridges a pre-trained waypoint predictor with
a geometry-aware local planner. Experiments in simulation and
on a physical humanoid robot show consistent gains in success
rate and path efficiency over contemporary baselines.
We position EgoNav as a systems contribution: rather than new
learning or planning algorithms, our value lies in identifying
why learned waypoint prediction fails in deployment and in the
module design and decision-layer coupling that resolve these
failures, as quantified by the progressive ablations.

Nonetheless, limitations remain. The waypoint predictor's generalization is bounded by its training distribution, and avoidance detours can drive the robot beyond the prior map's visual coverage, where the narrow field of view increases matching failures under viewpoint change. Panoramic map collection is a promising direction. A comparison against dedicated adaptive-planning schemes is another future direction.

\bibliographystyle{IEEEtran}
\bibliography{bible}

\begin{thebibliography}{10}
\providecommand{\url}[1]{#1}
\csname url@samestyle\endcsname
\providecommand{\newblock}{\relax}
\providecommand{\bibinfo}[2]{#2}
\providecommand{\BIBentrySTDinterwordspacing}{\spaceskip=0pt\relax}
\providecommand{\BIBentryALTinterwordstretchfactor}{4}
\providecommand{\BIBentryALTinterwordspacing}{\spaceskip=\fontdimen2\font plus
\BIBentryALTinterwordstretchfactor\fontdimen3\font minus \fontdimen4\font\relax}
\providecommand{\BIBforeignlanguage}[2]{{%
\expandafter\ifx\csname l@#1\endcsname\relax
\typeout{** WARNING: IEEEtran.bst: No hyphenation pattern has been}%
\typeout{** loaded for the language `#1'. Using the pattern for}%
\typeout{** the default language instead.}%
\else
\language=\csname l@#1\endcsname
\fi
#2}}
\providecommand{\BIBdecl}{\relax}
\BIBdecl

\bibitem{planner:fast-planner}
B.~Zhou, F.~Gao, L.~Wang, C.~Liu, and S.~Shen, ``Robust and efficient quadrotor trajectory generation for fast autonomous flight,'' \emph{IEEE Robotics and Automation Letters}, vol.~4, no.~4, pp. 3529--3536, 2019.

\bibitem{planner:ego-planner}
X.~Zhou, Z.~Wang, H.~Ye, C.~Xu, and F.~Gao, ``Ego-planner: An esdf-free gradient-based local planner for quadrotors,'' \emph{IEEE Robotics and Automation Letters}, vol.~6, no.~2, pp. 478--485, 2020.

\bibitem{poni}
S.~K. Ramakrishnan, D.~S. Chaplot, Z.~Al-Halah, J.~Malik, and K.~Grauman, ``Poni: Potential functions for objectgoal navigation with interaction-free learning,'' in \emph{Proceedings of the IEEE/CVF Conference on Computer Vision and Pattern Recognition}, 2022, pp. 18\,890--18\,900.

\bibitem{aligning}
N.~Xu, W.~Wang, R.~Yang, M.~Qin, Z.~Lin, W.~Song, C.~Zhang, J.~Gu, and C.~Li, ``Aligning knowledge graph with visual perception for object-goal navigation,'' in \emph{2024 IEEE International Conference on Robotics and Automation (ICRA)}.\hskip 1em plus 0.5em minus 0.4em\relax IEEE, 2024, pp. 5214--5220.

\bibitem{vln:navgpt}
G.~Zhou, Y.~Hong, and Q.~Wu, ``Navgpt: Explicit reasoning in vision-and-language navigation with large language models,'' in \emph{Proceedings of the AAAI Conference on Artificial Intelligence}, vol.~38, no.~7, 2024, pp. 7641--7649.

\bibitem{vln:towards}
X.~Song, W.~Chen, Y.~Liu, W.~Chen, G.~Li, and L.~Lin, ``Towards long-horizon vision-language navigation: Platform, benchmark and method,'' \emph{arXiv preprint arXiv:2412.09082}, 2024.

\bibitem{gnm}
D.~Shah, A.~Sridhar, A.~Bhorkar, N.~Hirose, and S.~Levine, ``Gnm: A general navigation model to drive any robot,'' in \emph{2023 IEEE International Conference on Robotics and Automation (ICRA)}.\hskip 1em plus 0.5em minus 0.4em\relax IEEE, 2023, pp. 7226--7233.

\bibitem{vint}
D.~Shah, A.~Sridhar, N.~Dashora, K.~Stachowicz, K.~Black, N.~Hirose, and S.~Levine, ``Vi{NT}: A foundation model for visual navigation,'' in \emph{Proceedings of the 7th Conference on Robot Learning (CoRL)}, ser. Proceedings of Machine Learning Research, vol. 229, 2023.

\bibitem{navigating}
J.~Krantz, T.~Gervet, K.~Yadav, A.~Wang, C.~Paxton, R.~Mottaghi, D.~Batra, J.~Malik, S.~Lee, and D.~S. Chaplot, ``Navigating to objects specified by images,'' in \emph{Proceedings of the IEEE/CVF International Conference on Computer Vision}, 2023, pp. 10\,916--10\,925.

\bibitem{viking}
D.~Shah and S.~Levine, ``Viking: Vision-based kilometer-scale navigation with geographic hints,'' \emph{arXiv preprint arXiv:2202.11271}, 2022.

\bibitem{deepexplore}
Y.~He, I.~Fang, Y.~Li, R.~B. Shah, and C.~Feng, ``Metric-free exploration for topological mapping by task and motion imitation in feature space,'' \emph{arXiv preprint arXiv:2303.09192}, 2023.

\bibitem{placenav}
L.~Suomela, J.~Kalliola, H.~Edelman, and J.-K. K{\"a}m{\"a}r{\"a}inen, ``Placenav: Topological navigation through place recognition,'' in \emph{2024 IEEE International Conference on Robotics and Automation (ICRA)}.\hskip 1em plus 0.5em minus 0.4em\relax IEEE, 2024, pp. 5205--5213.

\bibitem{habitat-web}
R.~Ramrakhya, E.~Undersander, D.~Batra, and A.~Das, ``Habitat-web: Learning embodied object-search strategies from human demonstrations at scale,'' in \emph{Proceedings of the IEEE/CVF conference on computer vision and pattern recognition}, 2022, pp. 5173--5183.

\bibitem{sem-object}
D.~S. Chaplot, D.~P. Gandhi, A.~Gupta, and R.~R. Salakhutdinov, ``Object goal navigation using goal-oriented semantic exploration,'' \emph{Advances in Neural Information Processing Systems}, vol.~33, pp. 4247--4258, 2020.

\bibitem{vlfm}
N.~Yokoyama, S.~Ha, D.~Batra, J.~Wang, and B.~Bucher, ``Vlfm: Vision-language frontier maps for zero-shot semantic navigation,'' in \emph{2024 IEEE International Conference on Robotics and Automation (ICRA)}.\hskip 1em plus 0.5em minus 0.4em\relax IEEE, 2024, pp. 42--48.

\bibitem{vlnsurvey}
J.~Gu, E.~Stefani, Q.~Wu, J.~Thomason, and X.~E. Wang, ``Vision-and-language navigation: A survey of tasks, methods, and future directions,'' \emph{arXiv preprint arXiv:2203.12667}, 2022.

\bibitem{mcgpt}
Z.~Zhan, L.~Yu, S.~Yu, and G.~Tan, ``Mc-gpt: Empowering vision-and-language navigation with memory map and reasoning chains,'' \emph{arXiv preprint arXiv:2405.10620}, 2024.

\bibitem{safevln}
L.~Yue, D.~Zhou, L.~Xie, F.~Zhang, Y.~Yan, and E.~Yin, ``Safe-vln: Collision avoidance for vision-and-language navigation of autonomous robots operating in continuous environments,'' \emph{IEEE Robotics and Automation Letters}, 2024.

\bibitem{msnav}
C.~Liu, Z.~Zhou, J.~Zhang, M.~Zhang, S.~Huang, and H.~Duan, ``Msnav: Zero-shot vision-and-language navigation with dynamic memory and llm spatial reasoning,'' \emph{arXiv preprint arXiv:2508.16654}, 2025.

\bibitem{nava3}
L.~Zhang, X.~Hao, Y.~Tang, H.~Fu, X.~Zheng, P.~Wang, Z.~Wang, W.~Ding, and S.~Zhang, ``$\mathrm{NavV}^3$: Understanding any instruction, navigating anywhere, finding anything,'' \emph{arXiv preprint arXiv:2508.04598}, 2025.

\bibitem{lp:simple}
N.~Buniyamin, W.~W. Ngah, N.~Sariff, Z.~Mohamad \emph{et~al.}, ``A simple local path planning algorithm for autonomous mobile robots,'' \emph{International journal of systems applications, Engineering \& development}, vol.~5, no.~2, pp. 151--159, 2011.

\bibitem{falco}
J.~Zhang, C.~Hu, R.~G. Chadha, and S.~Singh, ``Falco: Fast likelihood-based collision avoidance with extension to human-guided navigation,'' \emph{Journal of Field Robotics}, vol.~37, no.~8, pp. 1300--1313, 2020.

\bibitem{egoteb}
J.~S. Smith, R.~Xu, and P.~Vela, ``egoteb: Egocentric, perception space navigation using timed-elastic-bands,'' in \emph{2020 IEEE International Conference on Robotics and Automation (ICRA)}.\hskip 1em plus 0.5em minus 0.4em\relax IEEE, 2020, pp. 2703--2709.

\bibitem{lp:reshaping}
A.~Sarvesh, A.~Carroll, and S.~Gopalswamy, ``Reshaping local path planner,'' \emph{IEEE Robotics and Automation Letters}, vol.~7, no.~3, pp. 6534--6541, 2022.

\bibitem{planner:svsdf}
J.~Wang, T.~Zhang, Q.~Zhang, C.~Zeng, J.~Yu, C.~Xu, L.~Xu, and F.~Gao, ``Implicit swept volume sdf: Enabling continuous collision-free trajectory generation for arbitrary shapes,'' \emph{ACM Transactions on Graphics (TOG)}, vol.~43, no.~4, pp. 1--14, 2024.

\bibitem{iloc}
P.~Yin, S.~Zhao, J.~Wang, R.~Ge, J.~Ji, Y.~Hu, H.~Liu, and J.~Han, ``iloc: An adaptive, efficient, and robust visual localization system,'' \emph{IEEE Transactions on Robotics}, vol.~41, pp. 2709--2726, 2025.

\bibitem{mixvpr}
A.~Ali-Bey, B.~Chaib-Draa, and P.~Giguere, ``Mixvpr: Feature mixing for visual place recognition,'' in \emph{Proceedings of the IEEE/CVF winter conference on applications of computer vision}, 2023, pp. 2998--3007.

\bibitem{dformerv2}
B.-W. Yin, J.-L. Cao, M.-M. Cheng, and Q.~Hou, ``Dformerv2: Geometry self-attention for rgbd semantic segmentation,'' in \emph{Proceedings of the Computer Vision and Pattern Recognition Conference}, 2025, pp. 19\,345--19\,355.

\bibitem{habitat}
M.~Savva, A.~Kadian, O.~Maksymets, Y.~Zhao, E.~Wijmans, B.~Jain, J.~Straub, J.~Liu, V.~Koltun, J.~Malik \emph{et~al.}, ``Habitat: A platform for embodied ai research,'' in \emph{Proceedings of the IEEE/CVF international conference on computer vision}, 2019, pp. 9339--9347.

\bibitem{mp3d}
A.~Chang, A.~Dai, T.~Funkhouser, M.~Halber, M.~Niessner, M.~Savva, S.~Song, A.~Zeng, and Y.~Zhang, ``Matterport3d: Learning from rgb-d data in indoor environments,'' \emph{arXiv preprint arXiv:1709.06158}, 2017.

\bibitem{nomad}
A.~Sridhar, D.~Shah, C.~Glossop, and S.~Levine, ``Nomad: Goal masked diffusion policies for navigation and exploration,'' in \emph{2024 IEEE International Conference on Robotics and Automation (ICRA)}.\hskip 1em plus 0.5em minus 0.4em\relax IEEE, 2024, pp. 63--70.

\end{thebibliography}

\end{document}